\documentclass[11pt]{article}
\usepackage{paclic35}
\usepackage{times}
\usepackage{latexsym}
\usepackage{amsmath}
\usepackage{multirow}
\usepackage{url}

\usepackage{amsmath,bm}
\usepackage{lastpage}
\usepackage{url}
\usepackage{graphicx}
\graphicspath{{two_column_images/}}
\usepackage{array,multirow}
\usepackage{url}
\usepackage{bm}
\usepackage{soul}
\usepackage{amsmath}
\usepackage{color}
\usepackage[normalem]{ulem}
\usepackage{makecell}
\usepackage{upgreek}
\usepackage[ruled, lined, linesnumbered, commentsnumbered, longend]{algorithm2e}
\usepackage{xcolor}

\title{SEPP: Similarity Estimation of Predicted Probabilities\\for Defending and Detecting Adversarial Text}

\author{Hoang-Quoc Nguyen-Son, Seira Hidano, Kazuhide Fukushima, and Shinsaku Kiyomoto \\
  KDDI Research, Inc. \\
  2-1-15 Ohara, Fujimino, Saitama, 356-8502, Japan\\
  {\tt \{ho-nguyen,se-hidano,ka-fukushima,kiyomoto\}@kddi-research.jp} 
  }

\date{}

\begin{document}
\maketitle
\begin{abstract}
There are two cases describing how a classifier processes input text, namely, misclassification and correct classification.
In terms of misclassified texts, a classifier handles the texts
with both incorrect predictions and adversarial texts, which are generated to fool the classifier, which is called a victim.
Both types are misunderstood by the victim, but they can still be recognized by other classifiers.
This induces large gaps in predicted probabilities between the victim and the other classifiers.
In contrast, text correctly classified by the victim is often successfully predicted by the others and induces small gaps.
In this paper, we propose an ensemble model based on similarity estimation of predicted probabilities (SEPP) to exploit the large gaps in the misclassified predictions in contrast to small gaps in the correct classification.
SEPP then corrects the incorrect predictions of the misclassified texts.
We demonstrate the resilience of SEPP in defending and detecting adversarial texts through different types of victim classifiers, classification tasks, and adversarial attacks.

\end{abstract}

\section{Introduction}

Recent deep learning models have reached the human level in many NLP
tasks. 
However, these models are sensitive to changes in the input data.
An adversarial text can be generated from an original text while the original meaning is still preserved and bypasses human recognition.
However, adversarial text can fool many victims, such as sentiment analysis~\cite{ren2019generating}, question answering~\cite{jia2017adversarial}, and search engines~\cite{gil2019white}.

Popular adversarial text defenders are based on adversarial training~\cite{shrivastava2017learning,tramer2018ensemble} or modification detection~\cite{pruthi2019combating}.
$N$-gram~\cite{juuti2018stay} and text similarity~\cite{nguyen2019identifying} address the adversarial text detection problem.
However, recent generators can generate adversarial text via a very small change from the original by replacing a few words~\cite{ren2019generating,jin2020bert}, few characters~\cite{gao2018black,jones2020robust}, or both~\cite{li2019textbugger}.
High duplication in word usage between the original and adversarial texts confuses both the existing defenders and detectors.

\begin{figure*}[t]
\centering
\includegraphics[]{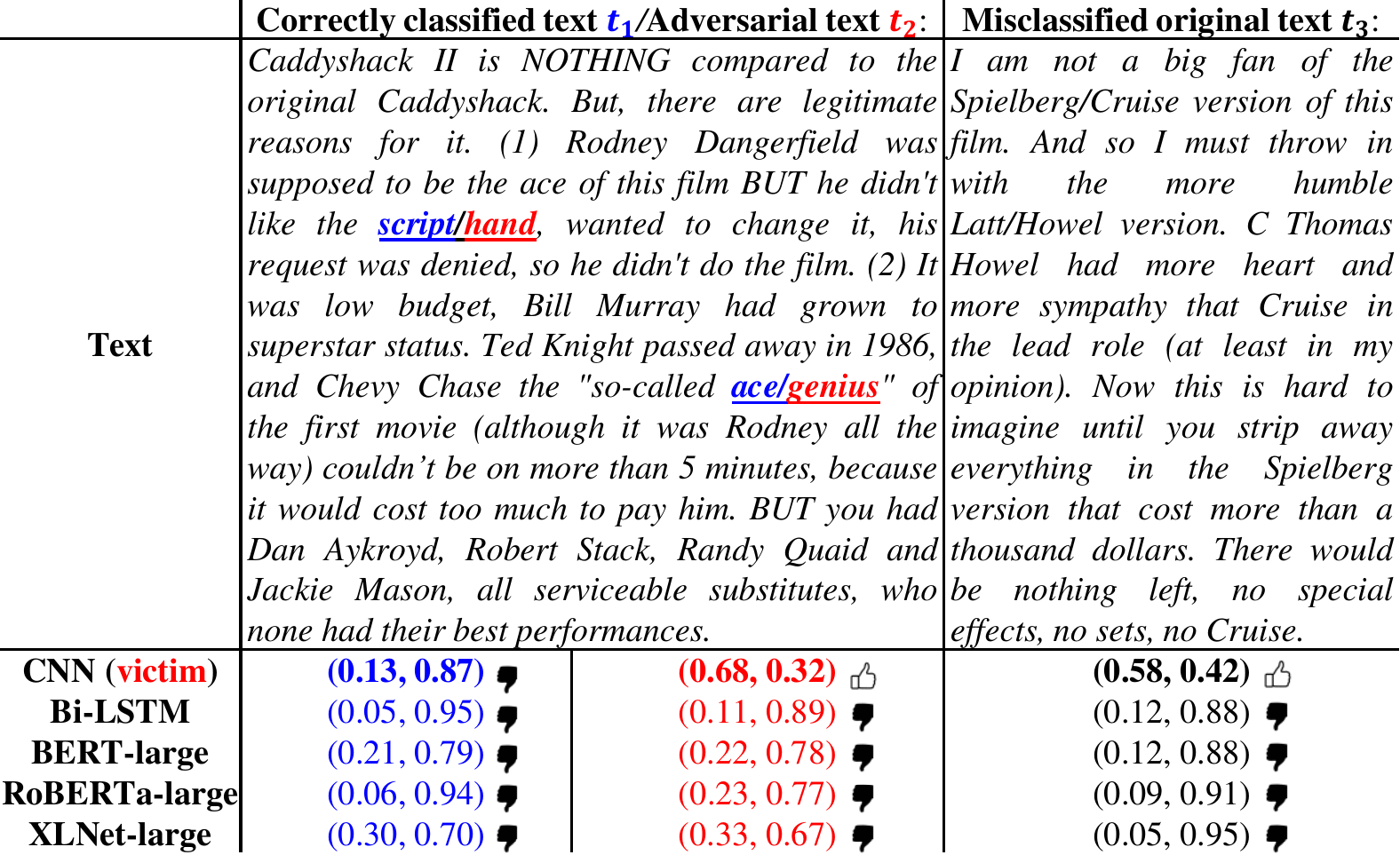}
\caption{Predictions (positive, negative) based on sentiment analysis classifiers.}
\label{Fig_1_Examples}
\end{figure*}

\subsection{Motivation} 
Correct classification of text by a classifier often induces small gaps to other classifiers.
An adversary can fool a victim classifier's predictions by generating misclassified text, but it does not fool other classifiers.
For instance, we randomly choose correctly classified text $t_1$ and its adversarial text $t_2$ targeting a CNN classifier (Figure~\ref{Fig_1_Examples}).
The predictions are made with popular deep learning models including CNN~\cite{kim2014convolutional}, BiLSTM, BERT-large~\cite{devlin2019bert}, RoBERTa-large~\cite{liu2019roberta}, and XLNet-large~\cite{yang2019xlnet}.
The prediction is indicated by pair values of positive and negative prediction probabilities.
The original text $t_1$ is negative, so the CNN and the other models correctly predict the text with higher negative than positive values.
The adversarial text $t_2$ changes two words ``\textit{script, ace}'' into their synonyms ``\textit{hand, genius}'' using ~\newcite{ren2019generating}'s work.
The generated text reduces the negative probability of the victim classifier to less than 0.5.
However, using the synonym does not change the overall meaning, so the other models mostly retain their negative predictions.
We randomly select a negative text $t_3$, which is misclassified by the CNN victim, and observe that $t_3$ has the same characteristic as $t_2$.
In particular, $t_3$ is predicted as positive by the victim, while other classifiers still predict it as negative.
Based on the gaps in prediction probabilities among the classifiers, we can distinguish correctly classified text from misclassified text.

\subsection{Contributions} 
In this paper, we proposed an ensemble model based on similarity estimation of predicted probabilities (SEPP) to defend adversarial texts.
Unlike a basic ensemble model, which directly votes predictions from multiple classifiers, SEPP estimates the similarity in prediction probabilities from the classifiers.
The similarity is used to identify the victim classifier and misclassified texts.
The probabilities of misclassified texts are corrected by using predictions from other classifiers.
We use the same technique to detect adversarial texts.

We conducted experiments with adversarial texts generated by the probability weighted word saliency  generator~\cite{ren2019generating} that fool the CNN-based sentiment analysis classifier. 
SEPP recovers the prediction from 22.9\% to 94.0\% on an adversarial dataset while keeping 96.6\% on the clean dataset.
This is better than the 89.6\% and 92.6\% achieved by adversarial training and ensemble baselines, respectively.
Moreover, we detect the adversarial texts at a rate of 96.3\%, which outperforms existing work, neural baselines, and ensemble baselines.
Other experiments on BiLSTM and BERT yield similar results.
SEPP also works well on multiple-class classification tasks and other adversarial attacks.
In summary, our contributions are as follows:

\begin{itemize}
\item 
We determined that predictions of various classifiers for misclassified text differ from those of correctly classified text.
\item
We proposed an ensemble model using similarity estimation of predicted probabilities (SEPP) to detect a victim classifier and misclassified texts.
We leveraged this detection to recover the prediction of the victim.
\item
We reused SEPP to distinguish adversarial text from the original text.
\item
We evaluated the various adversarial texts, which fooled CNN, BiLSTM, and BERT classifiers on binary- and multiple-class classification tasks.
The results indicate that SEPP outperforms other existing methods.
\end{itemize}

\subsection{Roadmap}
The rest of this paper is organized as follows. 
Section~\ref{section:related_work} describes related work on adversarial text generation, detection, and defense. 
Section~\ref{section:proposed_method} introduces the SEPP system. 
The experiential results are shown and analyzed in Section~\ref{section:evaluation}. 
Section~\ref{section:conclusion} summarizes some main key points and mentions future work. 

\section{Related Work}
\label{section:related_work}
\subsection{Adversarial Text Generation}

Adversarial text generation can be categorized by the extent of the generation:

\subsubsection{Paragraph} 
\newcite{juuti2018stay} trained a neural model on human-written reviews and generated adversarial texts by topic.
\newcite{jia2017adversarial} added a noise sentence to an original paragraph to change a correct result of a question answering system.
\newcite{wang2020cat} changed product categories of a review while keeping the sentiment but fooling a sentiment analysis classifier.

\subsubsection{Sentence}
\newcite{iyyer2018adversarial} generated an adversarial sentence with the desired syntax.
They used back-translation to create a paraphrased sentence pair with different syntax.
They then designed an attention network to convert a sentence into a paraphrase with the target syntax.
\newcite{ren2020generating} combined VAE and GAN to generate large scale adversarial sentences for a limited training dataset.
\newcite{han2020adversarial} generated a text using an RNN network targeting structured prediction models such as dependency parsing or POS tagger.

\subsubsection{Phrase}
\newcite{ribeiro2018semantically} compiled paraphrased pairs at the phrase level.
They then suggested a rule to replace individual phrases in an original text with corresponding phrases in the paraphrased pairs.
\newcite{liang2018deep} inserted or deleted consecutive hot words that affected the predictions of classifiers.
\newcite{wallace2019universal} added a fixed phrase at the beginning of any sentence and optimized it by the gradient of a victim system. 
They claim that a phrase ``zoning tapping fiences'' reduces the victim's accuracy from 86.2\% to 29.1\% on positive samples. 

\subsubsection{Word}
Adversarial text can be created by using various word operations (insertion, deletion, and replacement) to fool AI systems with both white-box and black-box attacks.
As an example of a white-box attack, \newcite{ebrahimi2018hotflip} operated on hot words that induce a high gradient change in the system.
As an example of a black-box attack, \newcite{liang2018deep} and \newcite{jin2020bert} examined occluded words and observed the prediction change.
\newcite{garg2020bae} marked candidate words and chose the top ones predicted by a BERT model.
\newcite{li2020bert} extended this idea for sub-words.
\newcite{zhang2019generating} improved the fluency of word replacement by performing Metropolis-Hastings sampling. 
The chance of replacement is improved by using a genetic algorithm~\cite{alzantot2018generating}, particle swarm optimization~\cite{zang2020word}, or boundary optimization~\cite{meng2020geometry}.
\newcite{ren2019generating} upgraded the text fluency with synonymous words in Wordnet and similar name entities.

\subsubsection{Character}
Many of the word-based approaches can be applied directly to characters~\cite{liang2018deep,ebrahimi2018hotflip}.
Moreover, \newcite{zhou2019learning} recovered the character replacement in an adversarial text.
\newcite{gil2019white} suggested a method based on a character operator targeting Google search scores.
\newcite{pruthi2019combating}, \newcite{jones2020robust}, and \newcite{li2019textbugger} manipulated the middle characters of an individual word to preserve the text fluency.

\paragraph{Analysis:} 
The \textit{paragraph} approach generates flexible adversarial texts. 
The generation of large hard-to-read text makes it easily recognizable by the $N$-gram model and readability metrics~\cite{juuti2018stay}.
The \textit{sentence} approaches preserves the text meaning, but they induce significant changes in text complexity~\cite{nguyen2019identifying}.
In the \textit{phrase} approach, the rules become fragile when we gather sufficient paraphrased pairs. 
The insertion and deletion of hot phrases into original text induces nonfluent text.
The operators on \textit{character} introduce misspellings.
With the \textit{word} operator, while insertion and deletion also lead to nonfluent text, the replacement produces fluent text.
Among these replacements, the Wordnet-based approach~\cite{ren2019generating} preserves more the original meaning than other replacements, which are based on word embedding~\cite{li2020bert,zang2020word}. 
Moreover, this replacement works well on many tasks (binary- or multiple-class classification) and is chosen to conduct main experiments in this paper.

\subsection{Adversarial Text Defense}

The most popular approach in the defense against adversarial text is adversarial training~\cite{shrivastava2017learning}, which was previously used in image processing.
The adversarial texts were added to the training data before the classifier was retrained.
Another approach estimated the similarity between original and adversarial texts on training data~\cite{liu2020joint}.
The upper and lower bounds of adversarial data were also approximated~\cite{ye2020safer,huang2019achieving,jia2019certified} to alleviate such texts.
Other defenses identified changes in adversarial texts from their origins at the character level~\cite{jones2020robust,pruthi2019combating} or word level~\cite{zhou2019learning}.
The main drawback of previous approaches is that they need to retrain the classifier.
Thus, they are sensitive to a new kind of adversarial text.

\subsection{Adversarial Text Detection}

Original text is generally more fluent than adversarial text.
Existing methods estimate the fluency based on the $N$-gram model.
\newcite{juuti2018stay} extracted the $N$-gram features based on a variety of text components, including word, part of speech, and syntactic dependency. 
They also measured the text readability using thirteen relative metrics.
Our previous work~\cite{nguyen2019identifying} extracted word $N$-gram features in both internal information from a training corpus and external information from a website corpus\footnote{\url{https://catalog.ldc.upenn.edu/LDC2006T13}}.
Text coherence was measured by matching similar words and combining them with the $N$-gram features.
Powerful deep learning models (e.g., BERT~\cite{devlin2019bert}, RoBERTa~\cite{liu2019roberta}, and XLNet~\cite{yang2019xlnet}) can be used as reputable detectors since they prove their performance in most of the major classification tasks.

Existing methods extract the difference in word usage between original and adversarial texts.
However, recent adversarial texts produced only minimal changes from the original texts.
Thus, they confuse all text-based methods.

\section{Similarity Estimation of Predicted Probabilities}
\label{section:proposed_method}

\begin{figure*}[t]
\centering
\includegraphics[]{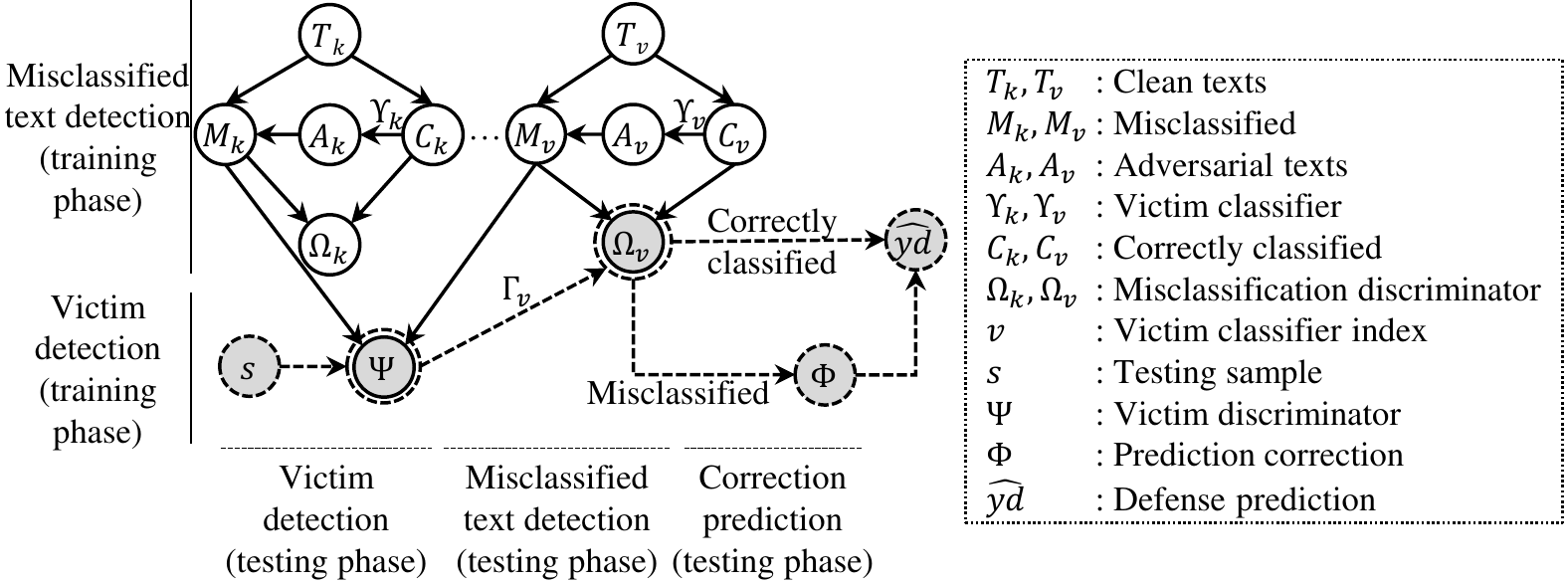}
\caption{Similarity estimation of the predicted probability of
  defending against adversarial text. Training and testing are shown as solid and dashed lines, respectively.}
\label{Fig_2_proposed_method}
\end{figure*}

We proposed an ensemble model based on the similarity estimation of predicted probabilities system (SEPP) for defending against adversarial text, as shown in Figure~\ref{Fig_2_proposed_method}.

\subsection{Training phase}

The objective of the training phase is to create two kinds of discriminators.
A discriminator $\Omega_k$ detects misclassified texts for a classifier $\Gamma_k$.
Another discriminator $\Psi$ detects a victim among candidate classifiers.

\subsubsection{Training Misclassification Discriminator $\bm{\Omega_k}$}

We describe the training of a misclassification discriminator $\Omega_k$ for a victim classifier $\Gamma_k$ in the following steps. The other misclassification discriminators is trained in the same manner way.

\begin{itemize}
    \item \textit{\textbf{Preparing training texts}}:
We run a victim classifier $\Upsilon_k$ to divide clean texts $T_k$ into misclassified texts $M_k$ and correctly classified texts $C_k$.
Adversarial texts $A_k$ are then generated from $C_k$ by using an existing generator and are added to $M_k$.
Each text $t$ in $M_k$ and $C_k$ is used to extract features for training $\Omega_k$ (Algorithm~\ref{alg:extract_features}).

\item\textit{\textbf{Measuring similarities:}}
The probability $\hat{y^c}$ of the predicted class $c$ in $\Upsilon_k$ is calculated with respect to its similarity to corresponding probabilities in other classifiers $\Gamma_i$.
In particular, the similarity is the Manhattan distance of $\hat{y^{c}}$ and $\hat{y^{c}_i}$ (line 7).

\item\textit{\textbf{Counting different predictions:}}
We count predicted classes of other classifiers $\Gamma_i$ that are different from the predicted class $c$ of the victim, which calls the different prediction count $\theta$ (line 9).

\item\textit{\textbf{Training the misclassification discriminator:}}
All similarities $\Lambda$ and $\theta$ are input into a feedforward neural network to train $\Omega_k$.

\end{itemize}

\newcommand\mycommfont[1]{\small\ttfamily\textcolor{black}{#1}}
\SetCommentSty{mycommfont}

\begin{algorithm}[t]
\SetKwFunction{getPredict}{getPredict}
\SetKwInOut{KwIn}{Input}
\SetKwInOut{KwOut}{Output}

\KwIn{text $t$;
victim $\Upsilon_k$;\\
other classifiers $\Gamma=\{\Gamma_i\}$}
\KwOut{extracted features}

$\hat{y}$ = \getPredict($\Upsilon_k,t$)

$c=\arg\max \hat{y}$

P = $\{\hat{y_i}=\getPredict(\Gamma_i,t)\}$

$\Lambda = \emptyset $ \tcp{similarity features}

$\theta =0$ \tcp{differences count feature}

\For{$\hat{y_i}\in P$} {
$\Lambda = \Lambda \cup |\hat{y^{c}} - \hat{y^{c}_i}| $

\If{$\arg\max \hat{y_i} \ne c$}{
$\theta = \theta + 1$
}
}
\KwRet{ $\Lambda \cup \theta$}
\caption{Extracting features.}
    \label{alg:extract_features}
\end{algorithm}

In Figure~\ref{Fig_1_Examples}, $t_1$, CNN, and other classifiers can be used as $t$, $\Upsilon_k$, and $\Gamma_i$, respectively.
$t_1$ is run on these classifiers to obtain $(positive, negative)$ probabilities $\hat{y} = (0.13, \bm{0.87}),\hat{y_1}=(0.05, \bm{0.95}), \cdots$.
The similarities are calculated as $\Lambda=(|\bm{0.87}-\bm{0.95}|=0.08,|\bm{0.87}-0.79|=0.08,\cdots)$.
All classifiers predict $t_1$ as negative; therefore, $\theta = 0$.
With the small values in $\Lambda$ and $\theta$, $t_1$ is most likely to be determined as correctly classified text.
With adversarial text $t_2$, the misclassified text should be detected with large values: $\Lambda=(0.57,\cdots), \theta=4$.
Similarly, $t_3$ should be considered misclassified text with $\Lambda=(0.56,\cdots), \theta=4$.

\subsubsection{Training Victim Discriminator $\bm{\Psi}$}
We use all misclassified texts to train a victim discriminator $\Psi$.
Each text extracts individual features from a victim classifier in the same manner as above.
The individual features are concatenated in order and input into another feedforward neural network to train $\Psi$.
When we use $t_2$ as the input text, individual features $(0.57, 0.46, ..., 4)$ and $(0.57, 0.11, ..., 1)...$ are extracted with $\Upsilon_1,\Upsilon_2...$.
The concatenated features $(0.57, 0.46, ..., 4,0.57, 0.11, ..., 1...)$ contain high values in the first individual features, so the first classifier should be identified.

\subsection{Testing phase}
A testing sample $s$ of adversarial or original text is run with $\Psi$ to determine the victim $\Upsilon_v$.
Then, the corresponding discriminator $\Omega_v$ determines whether $s$ is a correct or misclassified text.
If $\Omega_v$ determines $s$ as the correctly classified sample, then, we retain the original prediction on $\Upsilon_v$ for the final defense probability.
Otherwise, the defense probability is calculated by:
\begin{equation*}
\hat{yd} = \frac{1}{n}\sum_{i=1}^{n}{\hat{y_i}}
\end{equation*}
where $\hat{y_i}$ is the probability from other classifiers $\Gamma_i$ and $n$ is the total number of the other classifiers.

For example, if $\Psi$ identifies the victim $\Upsilon_1$ of
adversarial text $t_2$, $\Omega_1$ detects $t_2$ as misclassified
text. 
The prediction of $t_2$ is updated from positive with $\hat{y}=(\bm{0.58}, 0.42)$ to negative with:
$\hat{yd} = \left( \dfrac{0.11+0.22\cdots}{4}, \dfrac{0.89+0.78\cdots}{4} \right) 
= (0.22, \bm{0.78}).$
A similar flow should be processed with the misclassified text $t_3$.
In the case of the correctly classified text $t_1$, because this kind of text is already learned via all misclassification discriminators, the correctly classified text should be identified with any victim detected by $\Psi$.

\section{Evaluation}
\label{section:evaluation}

In this section, we present our experimental evaluation of defending and detecting adversarial texts.

\subsection{Defending against Adversarial Texts} 

\subsubsection{Dataset}
We created adversarial texts by using the probability weighted word saliency (PWWS) generator~\cite{ren2019generating} on the IMDB\footnote{\url{https://ai.stanford.edu/~amaas/data/sentiment/}} (binary class) and AGNEWS\footnote{\url{http://groups.di.unipi.it/~gulli/AG_corpus_of_news_articles.html}} (four classes).
We used testing data as a clean dataset.
The adversarial texts are replaced with the original texts from the clean dataset to form an adversarial dataset.
We use the ratio of $80/10/10$ for training/developing/testing sets.
This ratio is reused in further experiments.

\subsubsection{Comparison}
We compared SEPP\footnote{SEPP using five classifiers (Figure~\ref{Fig_1_Examples}), separately trained on the IMDB with suggested configurations and obtained similar performance.
For example, CNN and RoBERTa-large achieved $88.8\%$ and $96.5\%$ accuracies, respectively.} with adversarial training~\cite{shrivastava2017learning} and ensemble baselines~\cite{opitz1999popular}.
While adversarial training adds adversarial texts and retrains the
victims, ensemble learning votes on the predictions from the five individual classifiers (Figure~\ref{Fig_1_Examples}).
There are two popular ways to vote: average the predictions (soft) and select the majority class (hard).
Table~\ref{tab:defense_word_based} lists the accuracy scores on testing sets while the developing sets reach similar values.
SEPP can be trained with different training data (unknown), multiple training data (unsure), and the same training data (known).
For example, if a victim is CNN, the different (resp. multiple, same)
training data consists of misclassified and correctly classified
texts, $M_k$ and $C_k$ (Figure~\ref{Fig_2_proposed_method}), generated
with BiLSTM (resp. both BiLSTM and CNN, and CNN).

\begin{table*}[]
\centering
\begin{tabular}{l c c c c c c c c}
\hline
\multirow{3}{*}{\textbf{Method}} & \multicolumn{4}{c}{\textbf{IMDB}} & \multicolumn{4}{c}{\textbf{AGNEWS}}\\
& \multicolumn{2}{c}{\textbf{CNN}} & \multicolumn{2}{c}{\textbf{BiLSTM}} & \multicolumn{2}{c}{\textbf{CNN}} & \multicolumn{2}{c}{\textbf{BiLSTM}}\\
& \textbf{Clean} & \textbf{Adv} & \textbf{Clean} & \textbf{Adv} & \textbf{Clean} & \textbf{Adv} & \textbf{Clean} & \textbf{Adv}\\
\hline
Original (victim) & 88.9 & 22.9 & 87.0 & 14.2 & 91.7 & 62.6 & 92.2 & 56.5 \\
Adversarial training (known) & 88.4 & 89.6 & 86.0 & 86.6 & 92.2 & 88.2 & 91.6 & 85.3 \\
\hline
Ensemble (soft voting) & 95.0 & 92.6 & 95.0 & 92.6 & 94.7 & \underline{94.3} & 94.7 & \underline{94.2} \\
Ensemble (hard voting) & 96.0 & 90.6 & 96.0 & 91.3 & \underline{96.3} & 93.0 & \textbf{96.3} & 94.0 \\
\hline
SEPP (unknown) & \underline{96.3} & 90.9 & \underline{96.3} & 93.5 & 94.1 & 89.9 & 94.1 & 91.6 \\
SEPP (unsure) & \underline{96.3} & \textbf{94.8} & \underline{96.3} & \textbf{94.1} & 94.2 & 91.4 & 94.2 & 90.5 \\
SEPP (known) & \textbf{96.6} & \underline{94.0} & \textbf{96.6} & \textbf{94.1} & \textbf{96.5} & \textbf{95.7} & \underline{96.2} & \textbf{94.6} \\

\hline
\end{tabular}
\caption{Defending against adversarial texts targeting binary-class (IMDB) and multiclass (AGNEWS).}
\label{tab:defense_word_based}
\end{table*}

The victim classifier declines significantly when moving from clean to adversarial data.
Adversarial training efficiently defends against adversarial text, but it ignores the other misclassified texts.
Ensemble learning appropriates this task in which adversarial text fools the victim classifier only but the other classifiers are still persistent.
SEPP processes both kinds of misclassified texts and achieves high outcomes even with unknown victim classifiers.
Moreover, SEPP (unsure) detects the victim classifiers more than 90\% of accuracy.

\subsubsection{Ablation Studies}
We analyzed the contributions of the individual classifiers used in SEPP.
The victim CNN is combined with the individual classifiers (Table~\ref{tab:Combination}).
SEPP is presented with three groups of features: similarities $\Lambda$ (SEPP-$\Lambda$), differences $\theta$ (SEPP-$\theta$), and their combination (SEPP).
The detection is affected by the performance of each model.
In particular, BERT, RoBERTa, and XLNet are better than BiLSTM.
SEPP improves both predictions in individual and combined features.

\begin{table}[t]
\centering
\begin{tabular}{l c c c c}
\hline
\multirow{2}{*}{\textbf{Combination}} & \multicolumn{2}{c}{\textbf{IMDB}} & \multicolumn{2}{c}{\textbf{AGNEWS}}\\
& \textbf{Clean} & \textbf{Adv} & \textbf{Clean} & \textbf{Adv}\\
\hline
CNN+BiLSTM  & 87.3 & 68.0 & 92.2 & 83.0\\
CNN+BERT & 95.2 & 91.9 & 95.9 & 94.5\\
CNN+RoBERTa & \textbf{97.0} & \textbf{95.7} & 95.4 & 93.3 \\
CNN+XLNet & 95.0 & 92.6 & 94.7 & 94.3\\
\hline
SEPP-$\Lambda$ & 96.4 & 93.4 & 90.8 & 94.1 \\
SEPP-$\theta$ & \underline{96.6} &  \underline{94.1} & \textbf{96.6} & \underline{95.5} \\
SEPP (both) & \underline{96.6} & 94.0 & \underline{96.5} & \textbf{95.7} \\
\hline
\end{tabular}
\caption{Combination of classifiers and features in SEPP.}
\label{tab:Combination}
\end{table}

\subsubsection{Attacking the BERT}
We conducted other experiments (Table~\ref{tab:attacks_with_BERT}) targeting the BERT on SST-2 with various attacks at different text levels: character (DeepWordBug~\cite{gao2018black}), character and word (TextBugger~\cite{li2019textbugger}), and word (TextFooler~\cite{jin2020bert}).
We reused all six pretrained SST-2 classifiers for the ensemble models from the TextAttack framework~\cite{morris2020textattack} including CNN, LSTM, BERT-base, DistilBERT-base, RoBERTa-base, and AlBERT-base.
The change in a few SST-2 words ($8.7$ words/text) leads to a remarkable change in classifiers' predictions and negatively affects ensemble models, especially in hard voting.
However, SEPP retains the most efficient defenses across the attacks.

\begin{table*}[t]
\centering
\begin{tabular}{l c c c c}
\hline
\textbf{Method} & \textbf{Clean} & \textbf{DeepWordBug} & \textbf{TextBugger} & \textbf{TextFooler}\\
\hline
Victim & 88.6 & 10.2 & 14.8 & 2.27\\
Adversarial training & 88.6 & 45.5 & 67.4 & 54.5 \\
\hline
Ensemble (soft voting) & 87.5  & \underline{54.6} & \underline{69.3} & \underline{58.0} \\
Ensemble (hard voting) & \textbf{89.8} & 40.9 & 60.2 & 43.2  \\
\hline
SEPP & \textbf{89.8} & \textbf{55.7} & \textbf{72.7} & \textbf{63.4}  \\
\hline
\end{tabular}
\caption{Attacking BERT on SST-2.}
\label{tab:attacks_with_BERT}
\end{table*}

\subsection{Detecting Adversarial Texts}

\subsubsection{Detecting Adversarial Texts with Duplicate Replacement}

\begin{figure}[t]
\centering
\includegraphics[width=\columnwidth]{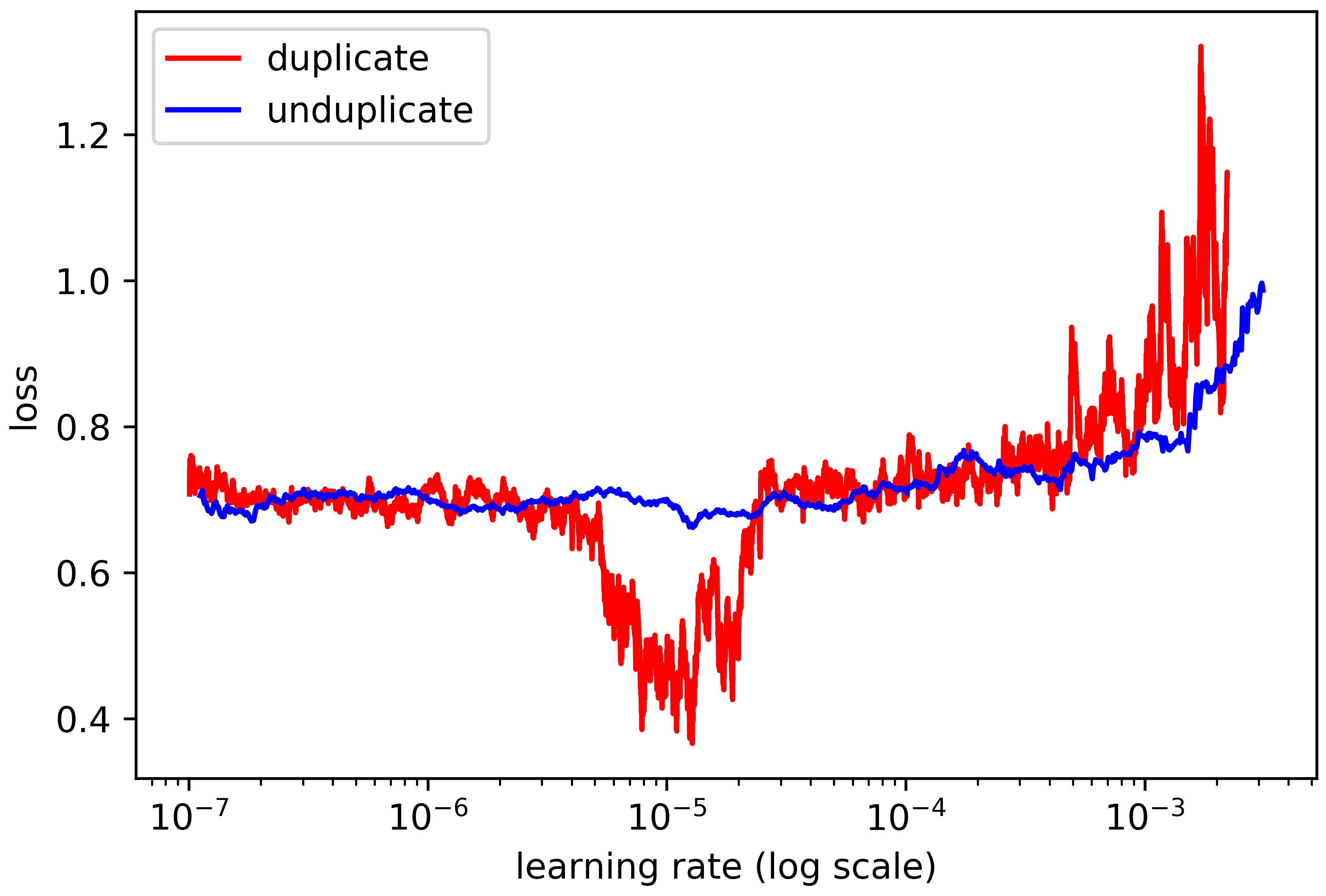}
\caption{Learning rate estimation of BERT-large model in duplicate and unduplicate replacement generation of adversarial texts.}
\label{fig:lr}
\end{figure}

We integrated adversarial texts with the original texts to form adversarial/original pairs.
These pairs are split into training/development/testing sets with the previous ratio (80/10/10).
SEPP detects adversarial texts by extracting the same kind of features as when detecting misclassified texts (see misclassification discriminator $\Omega_k$ in Figure~\ref{Fig_2_proposed_method}).
We compared SEPP with existing methods in detecting adversarial text, deep neural, and ensemble baselines as shown in Table~\ref{tab:adv_detection_all}.
The neural baselines were trained on large models with a batch size of 4, a maximum length of 512, and an epoch of 2.
The learning rates were estimated in a range of $10\mathrm{e}{-7}$ to $10\mathrm{e}{-2}$.
For example, Figure~\ref{fig:lr} shows the losses in the red line corresponding to the learning rates using the BERT-large model.
An optimal learning rate of $1.28\mathrm{e}{-5}$ was chosen when the loss was still decreasing, as recommended by \newcite{smith2017cyclical}.
The number of training/test sets is shown in the second row.

\begin{table*}[t]
\centering
\begin{tabular}{l c c c c}
\hline
\multirow{2}{*}{\textbf{Method}} & \multicolumn{2}{c}{\textbf{IMDB}} & \multicolumn{2}{c}{\textbf{AGNEWS}}\\
& \textbf{CNN} & \textbf{BiLSTM} & \textbf{CNN} & \textbf{BiLSTM}\\
\hline
\#train/\#test & 26531/3316 & 30046/3756 & 3376/422 & 2538/317 \\ 
\hline
$N$-gram
& 81.3 & 83.0 & 70.1 & 69.6\\
Complexity
& 80.0 & 82.9 & 73.2 & 68.7\\
BERT-large
& 92.7 & 91.5 & 89.3 & 88.7\\
RoBERTa-large
& \underline{95.0} & 94.9 & 88.4 & 93.8\\
XLNet-large
& 94.6 & 94.9 & 91.0 & 92.3\\
\hline
Ensemble (soft voting)
& 94.6  & \underline{96.7} & \textbf{94.8} & \textbf{97.0} \\
Ensemble (hard voting)
& 94.8 & 95.2 & 92.9 & 95.4\\
\hline
SEPP & \textbf{96.3} & \textbf{97.6} & \underline{94.3} & \underline{96.8}\\
\hline
\end{tabular}
\caption{Detecting adversarial texts with duplicate replacement.}
\label{tab:adv_detection_all}
\end{table*}

The results show that the deep neural and ensemble baselines efficiently enhanced the traditional approaches by more than 10\%.
SEPP\footnote{Our source code is available at the following link (\url{https://github.com/quocnsh/SEPP})} achieves the highest performances in binary-class classification algorithms and reaches the competitive performances in multi-class classification.

\subsubsection{Human Recognition}

We randomly chose 50 adversarial/original pairs in the development set for human recognition. 
They were shuffled, and each text was displayed to 11 raters who decided whether it was written by a human or generated by a machine\footnote{The survey is available at the following link (\url{https://forms.gle/TNRNeYyAcyrt8zF67})}. 
The raters recognized the adversarial texts with 62.1\% accuracy on average with a low agreement ($\kappa=-0.039$).
This recognition accuracy was lower than those of all machine detectors.
This demonstrates that we need a detector to assist us in recognizing such texts.

\subsubsection{Detecting Adversarial Texts with Unduplicated Replacement}

\begin{figure}[t]
\centering
\includegraphics[]{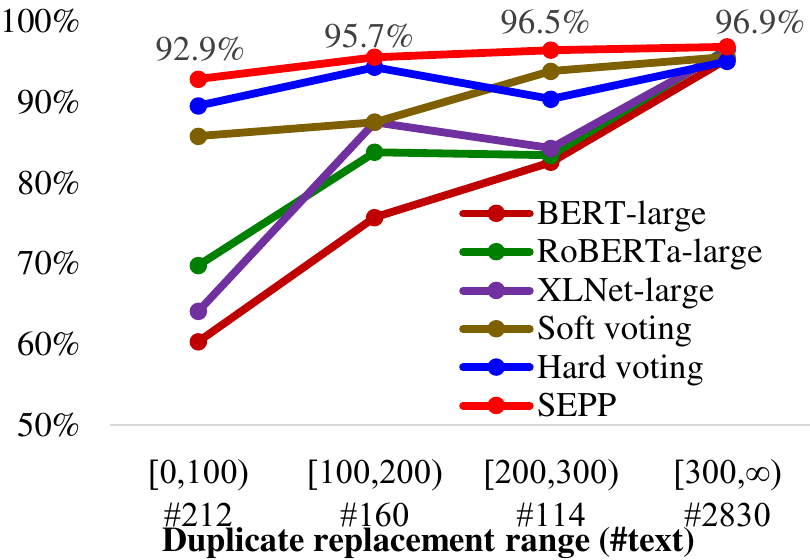}
\caption{Detection of adversarial texts that fool CNN classifier. 
Duplicate replacement indicates number of replacements reused in training data.}
\label{Fig:detection_by_relationships}
\end{figure}

We analyzed the PWWS generator and found that it uses a large number of duplicate word replacements to generate adversarial texts.
In particular, each replacement in a developing text was reused in 1544.3 texts on average in training texts. 
We clustered the texts in the development set in ranges of the number of duplicate replacements, as shown in Figure~\ref{Fig:detection_by_relationships}.
We compared the detection of the top six methods.
The low ranges significantly affected the deep learning baselines.
In the high ranges, many duplicate replacements occurred with training data, offering more chances for detection with these models.
However, since SEPP is independent of these replacements, we achieved resistant performances even in the low ranges.

\begin{table*}[t]
\centering
\begin{tabular}{l c c c c}
\hline
\multirow{2}{*}{\textbf{Method}} & \multicolumn{2}{c}{\textbf{IMDB}} & \multicolumn{2}{c}{\textbf{AGNEWS}}\\
& \textbf{CNN} & \textbf{BiLSTM} & \textbf{CNN} & \textbf{BiLSTM}\\
\hline
\#train & 1682 & 2028 & 972 & 1172 \\
\hline
$N$-gram
& 51.9 & 52.7 & 55.6 & 56.8\\
Complexity
& 51.1 & 51.2 & 50.8 & 53.4\\
BERT-large
& 50.9 & 51.6 & 56.5 & 63.5\\
RoBERTa-large
& 50.0 & 54.0 & 52.4 & 50.0\\
XLNet-large
& 50.0 & 55.9 & 50.0 & 62.2\\
\hline
Ensemble (soft voting)
& \underline{89.1} & 88.6 & 91.9 & \textbf{96.6} \\
Ensemble (hard voting)
& \underline{89.1} & \underline{89.0} & \textbf{94.4} & 95.2 \\
\hline
SEPP & \textbf{89.6} & \textbf{89.8} & \underline{92.7} & \underline{95.9}\\
\hline
\end{tabular}
\caption{Detecting adversarial texts with unduplicated replacement.}
\label{tab:adv_detection_unduplicate}
\end{table*}

We used PWWS to generate adversarial texts without reusing previous word replacements. 
We ran the detectors on this dataset (Table~\ref{tab:adv_detection_unduplicate}).
While existing methods and deep neural baselines remained in the random guess range, SEPP and ensemble baselines accuracy also maintained the prediction at around 92\%.
We analyzed the learning rate estimation process of the BERT-large model, as shown by the blue line (Figure~\ref{fig:lr}).
All of the losses were similar to a random line ($-ln(0.5)=0.69$).
The losses remained after many epochs of training.

\section{Conclusion}
\label{section:conclusion}

In this paper, we propose an ensemble model based on similarity
estimation of predicted probabilities (SEPP) for defending against adversarial text by detecting a victim classifier and correcting misclassified text.
SEPP measures the similarity among predictions from multiple classifiers.
We evaluated adversarial texts generated by word-based and/or character-based generators.
The generated texts targeted popular classifiers (CNN, BiLSTM, and BERT) in a binary and a multiclass classification.
The results show that SEPP outperformed the existing work not only in defending against adversarial texts but also in maintaining performance on clean texts.
Moreover, we achieved better performance in detecting adversarial texts than existing detectors.

Based on the generalization of the proposed method, we can straightforwardly apply it for detecting other adversarial data such as fake images or forged audio.

\bibliography{paclic35}
\bibliographystyle{acl}
\end{document}